\documentclass[runningheads]{llncs}
\usepackage[T1]{fontenc}
\usepackage{graphicx}

\usepackage{lipsum}  
\usepackage{amsmath}
\usepackage{multirow}
\usepackage{adjustbox}
\usepackage[dvipsnames]{xcolor}
\usepackage{url}

\begin{document}
\title{Unsupervised Training of Neural Cellular Automata on Edge Devices}
%
%
\author{John Kalkhof\orcidID{0000-0001-7316-1903}\thanks{Corresponding author: john.kalkhof@gris.tu-darmstadt.de, +49 6151 155-681} \and
Amin Ranem\orcidID{0000-0003-0783-6903} \and
Anirban Mukhopadhyay\orcidID{0000-0003-0669-4018}}
\authorrunning{J.Kalkhof et al.}

\institute{Darmstadt University of Technology, Karolinenplatz 5, 64289 Darmstadt, Germany}

\maketitle              
\begin{abstract}
The disparity in access to machine learning tools for medical imaging across different regions significantly limits the potential for universal healthcare innovation, particularly in remote areas. Our research addresses this issue by implementing Neural Cellular Automata (NCA) \textbf{training directly on smartphones} for accessible \emph{X-ray lung segmentation}. We confirm the practicality and feasibility of deploying and training these advanced models on five Android devices, improving medical diagnostics accessibility and bridging the tech divide to extend machine learning benefits in medical imaging to low- and middle-income countries (LMICs). We further enhance this approach with an unsupervised adaptation method using the novel \emph{Variance-Weighted Segmentation Loss (VWSL)}, which efficiently learns from unlabeled data by minimizing the variance from multiple NCA predictions. This strategy notably improves model adaptability and performance across diverse medical imaging contexts without the need for extensive computational resources or labeled datasets, effectively lowering the participation threshold. Our methodology, tested on three multisite X-ray datasets—Padchest, ChestX-ray8, and MIMIC-III—demonstrates improvements in segmentation Dice accuracy by 0.7 to 2.8\%, compared to the classic Med-NCA. Additionally, in extreme cases where no digital copy is available and images must be captured by a phone from an X-ray lightbox or monitor, VWSL enhances Dice accuracy by 5-20\%, demonstrating the method's robustness even with suboptimal image sources.

\end{abstract}
\section{Introduction}
Global access to advanced machine-learning tools for medical imaging reveals a significant divide between well-resourced environments and remote, underserved areas. This gap stems not just from the availability of technology but also from the infrastructure demands of conventional deep learning architectures, such as UNets \cite{UNet} and Transformers \cite{vaswani2017attention}. To address this issue, it is essential to focus on frugal healthcare solutions tailored for the last billion, emphasizing lightweight, efficient models and deployment strategies leveraging existing mobile technology \cite{mccool2022mobile}. These strategies aim to \emph{extend the reach of advanced diagnostics to underserved populations}. Conventional models, effective for medical image segmentation, require significant computational resources for training and deployment~\cite{kalkhof2023med}. Such demands restrict their use to well-equipped environments, limiting their adaptability to local data variations \cite{adam2024medical} in LMICs \cite{resourceConstrained_overview,frija2021improve}. While there are minimal UNets optimized for fewer parameters, they cannot match the performance of their more complex UNet counterparts \cite{kalkhof2023med}. This discrepancy highlights a significant obstacle to universal healthcare innovation and the provision of personalized care for diverse global populations.

\begin{figure}[t]
  \centering
  \includegraphics[width=.97\linewidth]{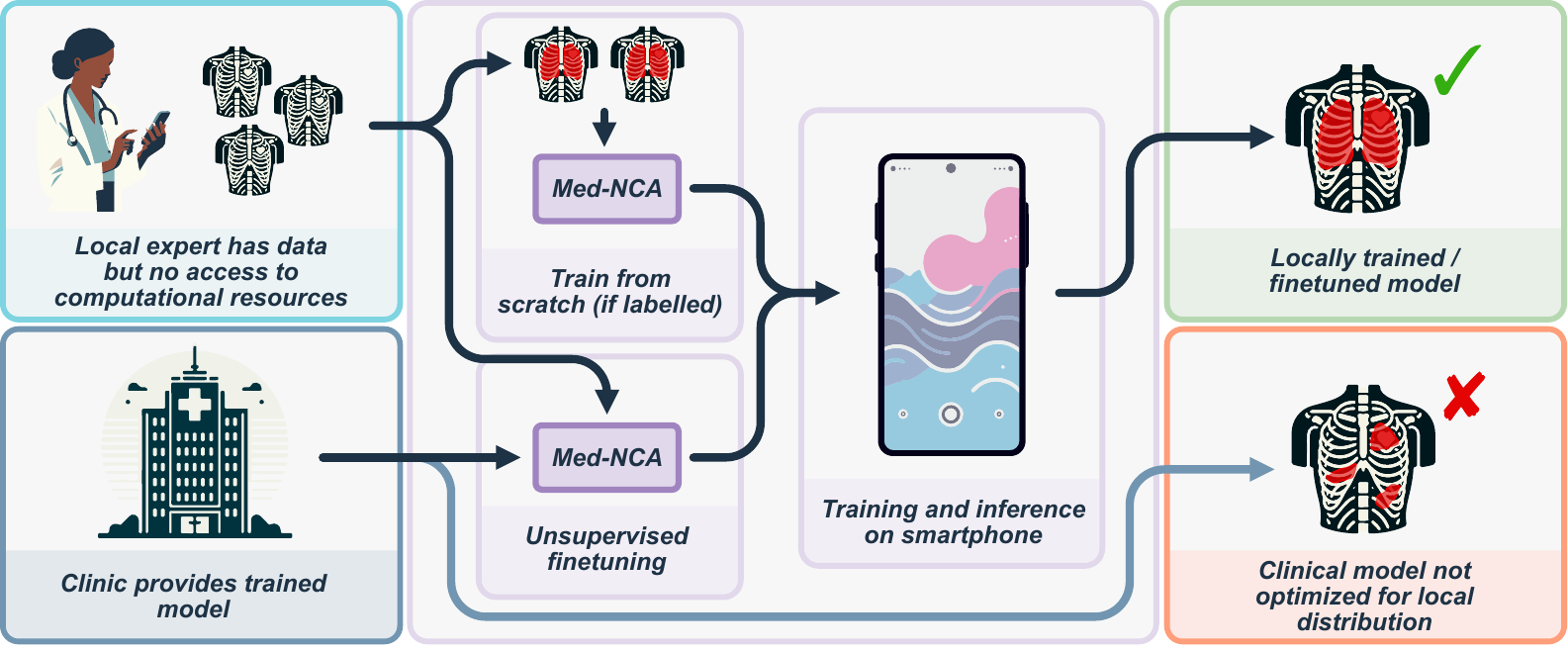}
  \caption{Neural Cellular Automata enable primary care professionals with data but no access to infrastructure, to train their models or fine-tune a pre-trained model on a smartphone.}
  \label{fig:IntroductionGraphic}
\end{figure}

In contrast, Neural Cellular Automata (NCA) \cite{gilpin2019cellular,growingNCA} presents a compelling alternative to transcend these limitations. Their lightweight nature and efficient processing capabilities make NCAs uniquely suited for deployment on widely available devices, such as smartphones, thereby democratizing access to advanced diagnostic tools. Moreover, NCAs are \emph{inherently adaptable} and capable of evolving in response to data variations without the need for extensive computational resources \cite{kalkhof2023med}. This adaptability is crucial for personalizing care (Figure \ref{fig:IntroductionGraphic}), as it allows NCAs to be \emph{fine-tuned on local datasets in remote areas}, lowering the barrier towards geographical and demographic diversity in medical data.

Our work's major technical contribution is a novel unsupervised adaptation method for NCAs, the \textbf{Variance}\textbf{-Weighted Segmentation Loss (VWSL)}, that leverages their adaptability to optimize performance on unlabelled data. By carefully balancing the fine-tuning of model parameters and introducing controlled variance, our method enables NCAs to learn effectively from diverse data sources. This approach enhances model performance against distribution shifts caused by heterogeneous data sources, enabling NCAs to \emph{adapt to local data without extensive infrastructure or labeled data}—critical for LMICs where conventional models like UNets are not feasible due to limited computational resources.

In our comparison with traditional UNet architectures \cite{chen2021transunet}, advanced models such as nnUNet \cite{nnUnet} and Trans UNet \cite{chen2021transunet}, and the Med-NCA \cite{kalkhof2023med} model, we underscore the effectiveness of our VWSL method in \emph{segmenting the right and left lung in X-ray images}. X-rays are a widely available and cost-effective imaging modality, making them crucial for diagnosing and managing lung conditions in diverse healthcare settings. We selected Padchest \cite{bustos2020padchest}, ChestX-ray8~\cite{wang2017chestx}, and MIMIC-III \cite{johnson2016mimic} datasets, using ChexMask masks \cite{gaggion2023chexmask} for segmentation, to mirror the variety of clinical settings and underscore the adaptability of our approach. Our method not only improves upon Med-NCA results by 0.7 to 2.8\% in Dice score, but it often surpasses standard UNet models and also shows competitive performance with the nnUNet. Additionally, our experiments demonstrate our VWSL's capability to enhance segmentation Dice accuracy by 5-20\% when capturing X-ray images directly from monitors, showcasing adaptability even in unconventional imaging setups. Crucially, \emph{we demonstrate that NCA-based methods can be efficiently trained and fine-tuned directly on smartphones}, highlighting a significant advancement towards making personalized, accessible, and decentralized diagnostic solutions feasible in resource-limited settings. We make our whole framework available under \url{github.com/MECLabTUDA/M3D-NCA}.

\section{Methodology}

NCAs represent a novel class of models characterized by their dynamic and adaptable nature, driven by \emph{probabilistic updates}. This stochasticity enables NCAs to produce diverse predictions upon multiple executions, a feature not commonly found in traditional neural networks. The capacity for NCAs to yield different outcomes from the same input, due to their inherent stochasticity, has been explored in prior research primarily to gauge the reliability of predictions \cite{kalkhof2023m3d}. Unlike these previous applications, our approach harnesses the variability inherent in NCAs to enhance model training and adaptation to new domains.

\subsection{Adapting to New Domains with Variance-Based Fine-Tuning}

We train Med-NCA for 1500 epochs, following the specifications in the original publication with 16 channels, and a batch normalization layer \cite{kalkhof2023med,kalkhof2023m3d}, on a dataset with segmentation labels, typically sourced from a clinical setting. This initial training phase equips the Med-NCA model with the ability to perform segmentation tasks relevant to the medical domain of interest. Upon completion, the trained Med-NCA model, due to its minimal size of \emph{only 110.6\:kB}, can be distributed to remote locations through various means, including direct phone-to-phone transfer, email, or integrated within instant messaging platforms, ensuring deployment versatility. In these remote locations, data is available, yet a significant challenge emerges from the absence of labels.

\begin{figure}[t]
  \centering
  \includegraphics[width=.97\linewidth]{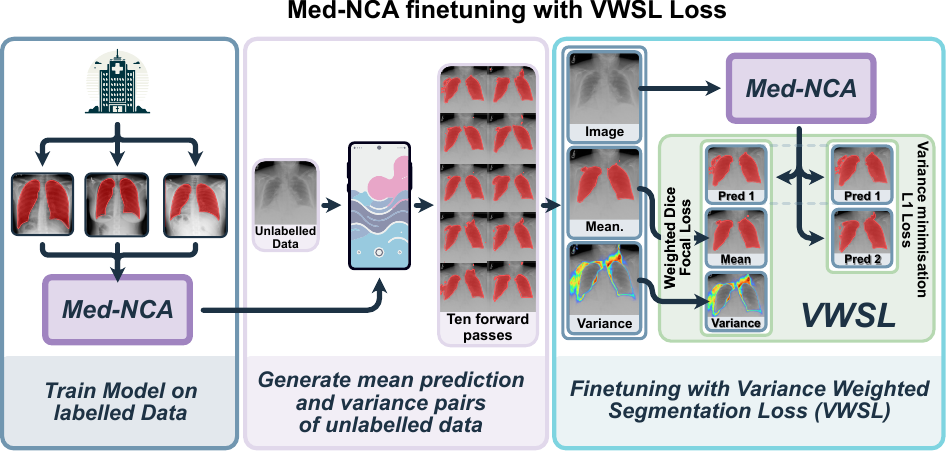}
  \caption{Our variance-based fine-tuning VWSL leverages mean predictions from the pre-trained Med-NCA, along with the corresponding variances. This process enables us to refine the model by targeting these predictions and modulating their impact based on variance. Such an approach ensures the model's adaptation to new domains while preserving crucial information.}
  \label{fig:Method}
\end{figure}

\begin{figure}[htbp]
  \centering
  \includegraphics[width=.85\linewidth]{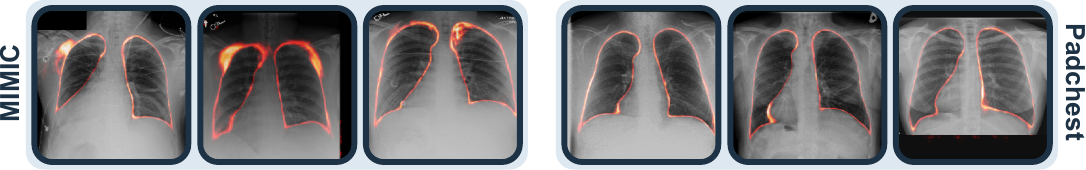}
  \caption{Variance of Med-NCA trained on ChestX-ray8 and evaluated on MIMIC and Padchest for segmentation of both lungs.}
  \label{fig:variance_map}
\end{figure}

To address this challenge, our novel approach involves a \emph{two-phase procedure} that utilizes the variability in Med-NCA's outputs. This process is detailed in Figure \ref{fig:Method}, illustrating the method's application. Initially, we compute the mean and variance pairs of ten predictions from the Med-NCA on unlabeled data at the remote location. We use the \emph{mean prediction as a surrogate target} for the subsequent fine-tuning phase, directing the model towards the intended segmentation results, preventing the trivial solution of setting all outputs to zero.  Simultaneously, we derive variance maps from these predictions, specifically calculating the standard deviation at each pixel. These maps \emph{illuminate regions of high variance} in the predictions (illustrated in Figure \ref{fig:variance_map}), indicating areas where the model's output fluctuates significantly. The objective during fine-tuning is to minimize this variance, thereby enhancing the model's consistency and accuracy in its predictions.

\subsection{Variance-Weighted Segmentation Loss (VWSL)}
At the heart of our approach is the novel \textbf{Variance-Weighted Segmentation Loss (VWSL)}, designed to refine the adaptation of Neural Cellular Automata (NCAs) for medical image segmentation. The VWSL integrates the Dice similarity coefficient (DSC) for overlap accuracy and a modified Focal Loss \cite{lin2017focal}, essential for emphasizing hard-to-segment areas, both adjusted by \emph{pixel-wise variance weighting} as the surrogate target. In addition, for each forward pass, two segmentation outputs $o_1$ and $o_2$ are generated per level, and their difference is minimized. This integrated loss function is formulated as follows:

\[ \text{VWSL} = w(x, y) \cdot [(1 - \text{DSC}) + \text{FocalLoss}] + \gamma * L1(o_1, o_2) \]  

where \(w(x, y) = (1 - 2*\text{Var}(x, y))\) represents the variance-based weighting for each pixel at position \((x, y)\), with \(\text{Var}(x, y)\) indicating the variance across predictions. The regulation factor \(\gamma\) balances the contribution of the variance minimization loss within the overall loss calculation.

The VWSL strategy directs the training for 100 additional epochs to align predictions with target segmentation while adaptively modulating learning based on the confidence level suggested by prediction variance. The VWSL thus enables targeted model fine-tuning, leveraging the inherent stochasticity of NCAs for robust adaptation to new domains without labeled data.

\section{Experimental Results}

Our evaluation of the VWSL fine-tuning loss, Med-NCA, and UNet-type methods focuses on X-ray segmentation of both lungs, utilizing three key datasets: Padchest \cite{bustos2020padchest}, ChestX-ray8 \cite{wang2017chestx}, and MIMIC-III \cite{johnson2016mimic}, with ChexMask \cite{gaggion2023chexmask} providing the according segmentation masks. Emulating low-resource settings, we scale images to $256 \times 256$ and utilize \emph{distinct sets of 50 samples} from each dataset for training, validation, and testing, respectively. This setup aims to replicate scenarios of limited data availability. To ensure consistency in our comparisons, UNet \cite{fernandoGarcia}, TransUNet, and Med-NCA were all implemented using PyTorch \cite{paszke2019pytorch} within the same framework, with the exception of nnUNet, which operates as a comprehensive end-to-end auto ML pipeline. All experiments were conducted on an \emph{Nvidia RTX 3090Ti} and an \emph{Intel Core i7-12700}. In addition, we also measure the training time across five different Android devices, namely the \emph{Moto G31}, \emph{Pixel 1 XL}, \emph{Poco M5}, \emph{Samsung S10} and the \emph{Pixel 6a}.

\begin{table}
\centering
\begin{adjustbox}{width=\textwidth,center}
\begin{tabular}{|c|c|c|c|c|c|c|} 
\hline
 \textbf{Train} & \multirow{2}{*}{\textbf{Method}} & \multicolumn{3}{c|}{\textbf{Dice} $\uparrow$ $\pm$ \textbf{STD} $\downarrow$} & \multirow{2}{*}{\textbf{Parameters} $\downarrow$}  & \textbf{Runs on}\\ 
 \textbf{On} & & \textbf{MIMIC}  & \textbf{ChestX8} & \textbf{Padchest} & & \textbf{Phone}\\
\hline
\multirow{5}{*}{\rotatebox[origin=c]{45}{\textbf{MIMIC}}} & Med-NCA & 0.796 $\pm$ 0.123 & 0.816 $\pm$ 0.080 & 0.849 $\pm$ 0.052 & 26432 & \textcolor{Green}{Yes}\\
 & VWSL Finetuning & - & 0.830 $\pm$ 0.068 & 0.867 $\pm$ 0.043 & 26432 & \textcolor{Green}{Yes}\\
 & UNet & 0.815 $\pm$ 0.113 & 0.792 $\pm$ 0.104 & 0.866 $\pm$ 0.045 & 36950273 & \textcolor{Red}{No}\\
 & TransUNet & 0.836 $\pm$ 0.118 & 0.858 $\pm$ 0.065 & 0.898 $\pm$ 0.031 & 105323161 & \textcolor{Red}{No}\\
 & nnUNet & 0.849 $\pm$ 0.132 & 0.828 $\pm$ 0.098 & 0.853 $\pm$ 0.126 & 29966112 & \textcolor{Red}{No}\\
\hline
\multirow{5}{*}{\rotatebox[origin=c]{45}{\textbf{ChestX8}}} & Med-NCA & 0.756 $\pm$ 0.132 & 0.899 $\pm$ 0.107 & 0.941 $\pm$ 0.061 & 26432 & \textcolor{Green}{Yes}\\
 & VWSL Finetuning & 0.784 $\pm$ 0.124 &  - & 0.955 $\pm$ 0.033 & 26432 & \textcolor{Green}{Yes}\\
 & UNet & 0.770 $\pm$ 0.128 & 0.887 $\pm$ 0.108 & 0.958 $\pm$ 0.020 & 36950273 & \textcolor{Red}{No}\\
 & TransUNet & 0.793 $\pm$ 0.119 & 0.929 $\pm$ 0.078 & 0.969 $\pm$ 0.020 & 105323161 & \textcolor{Red}{No}\\
 & nnUNet & 0.779 $\pm$ 0.149 & 0.914 $\pm$ 0.106 & 0.949 $\pm$ 0.135 & 29966112 & \textcolor{Red}{No}\\
\hline
\multirow{5}{*}{\rotatebox[origin=c]{45}{\textbf{Padchest}}} & Med-NCA & 0.762 $\pm$ 0.135 & 0.877 $\pm$ 0.134 & 0.954 $\pm$ 0.052 & 26432 & \textcolor{Green}{Yes}\\
 & VWSL Finetuning & 0.775 $\pm$ 0.138 & 0.884 $\pm$ 0.117 & - & 26432 & \textcolor{Green}{Yes}\\
 & UNet & 0.782 $\pm$ 0.132 & 0.872 $\pm$ 0.118 & 0.957 $\pm$ 0.028 & 36950273 & \textcolor{Red}{No}\\
 & TransUNet & 0.796 $\pm$ 0.122 & 0.925 $\pm$ 0.076 & 0.972 $\pm$ 0.018 & 105323161 & \textcolor{Red}{No}\\
 & nnUNet & 0.795 $\pm$ 0.128 & 0.925 $\pm$ 0.085 & 0.962 $\pm$ 0.091 & 29966112 & \textcolor{Red}{No}\\
\hline

\hline
\end{tabular}
\end{adjustbox}
\newline
\caption{Comparison of generalization performance across different datasets of Med-NCA, fine-tuned Med-NCA \emph{('-' denotes same domain)}, and UNet type baselines.}
\label{Tab:QuantHP}
\end{table}

\subsection{Quantitative Comparison and Ablation}
The results presented in Table \ref{Tab:QuantHP} underscore the effectiveness of fine-tuning Med-NCA across all initial datasets, crucial in real-world scenarios where the training dataset is determined by availability. They demonstrate notable improvements in Dice scores across all datasets when compared to the original Med-NCA and even challenge more parameter-heavy models such as the UNet and nnUNet. Specifically, the fine-tuned Med-NCA shows improvements ranging from 0.7 to 2.8\% in Dice scores across the datasets, indicating a substantial enhancement in generalization performance to new, unseen domains \emph{without requiring labels for fine-tuning}.

Moreover, the fact that both the original and fine-tuned versions of Med-NCA can run on smartphones, as evidenced by successful deployments on various devices, contrasts sharply with the high-parameter models that are not feasible for mobile execution. This distinction highlights the practical benefits of Med-NCA, offering a balance between model complexity, performance, and deployability in real-world, resource-constrained settings.

\textbf{Ablation: } The ablation study, detailed in Table \ref{Tab:Ablation}, explores the impact of varying the \(\gamma\) weighting in the Variance-Weighted Segmentation Loss (VWSL) on model performance. As \(\gamma\) increases from 0 to \(10^3\), we observe an improvement in Dice scores, \emph{peaking at \(\gamma = 10^3\)} with scores of 0.775 $\pm$ 0.138 on MIMIC and 0.884 $\pm$ 0.117 on ChestX8. However, further increasing \(\gamma\) to \(10^4\) slightly reduces performance, suggesting an optimal range for \(\gamma\) that balances variance influence without overshadowing other loss components. This nuanced optimization demonstrates the crucial role of \(\gamma\) in enhancing the adaptability and accuracy of the fine-tuned Med-NCA, especially in settings with limited training data.

\begin{table}[b]
\centering
\begin{adjustbox}{center}
\begin{tabular}{|c|c|c|c|} 
\hline
\textbf{Train} & \multirow{2}{*}{\textbf{Method}} & \multicolumn{2}{c|}{\textbf{Dice} $\uparrow$ $\pm$ \textbf{STD} $\downarrow$} \\
\textbf{On} & & \textbf{MIMIC} & \textbf{ChestX8} \\ 
\hline
\multirow{4}{*}{\rotatebox[origin=c]{45}{\textbf{Padchest}}} & VWSL \(\gamma\) = 0 & 0.753 $\pm$ 0.141 & 0.866 $\pm$ 0.140 \\
& VWSL \(\gamma\) = $10^2$ &  0.758 $\pm$ 0.142  &  0.884 $\pm$ 0.122 \\
& VWSL \(\gamma\) = $10^3$ &  0.775 $\pm$ 0.138 & 0.884 $\pm$ 0.117  \\
& VWSL \(\gamma\) = $10^4$ &  0.761 $\pm$ 0.147 &  0.871 $\pm$ 0.122 \\
\hline
\end{tabular}
\end{adjustbox}
\newline
\caption{Influence of \(\gamma\) weighting on the variance-based optimization performance.}
\label{Tab:Ablation}
\end{table}

\subsection{Qualitative Comparison}
The qualitative comparison in Figure \ref{fig:variance_map} illustrates the transformations achieved by fine-tuning Med-NCA with the Variance-Weighted Segmentation Loss (VWSL). Initially, the segmentations are marked by noticeable errors, ranging from significant over-segmentation to the omission of crucial regions. Yet, the application of VWSL for fine-tuning leads to considerable improvements, yielding segmentations that are noticeably more accurate. This comparison demonstrates the potential of the VWSL to refine suboptimal initial segmentations into much more precise results. 

\begin{figure}[t]
  \centering
  \includegraphics[width=.95\linewidth]{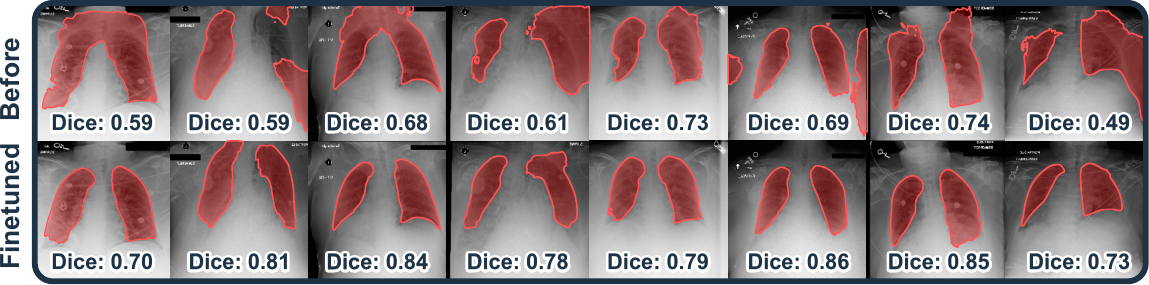}
  \caption{Med-NCA trained on the ChestX8 dataset before and after being fine-tuned on the MIMIC dataset.}
  \label{fig:finetuning_results}
\end{figure}

\subsection{VWSL Improvement on Smartphone X-ray Images}

\begin{figure}[b]
  \centering
  \includegraphics[width=.9\linewidth]{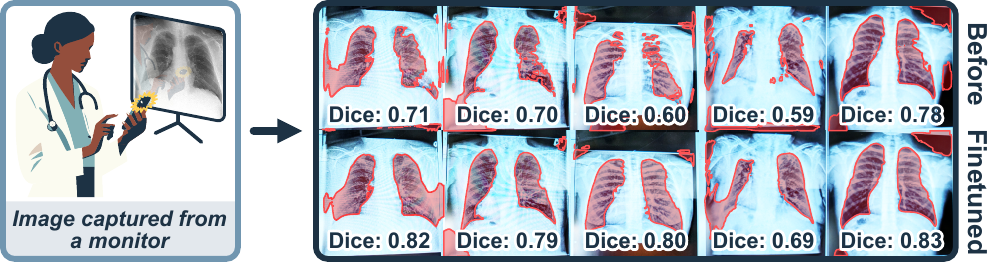}
  \caption{Med-NCA trained on the Padchest dataset before and after being fine-tuned on pictures taken by a Pixel XL.}
  \label{fig:monitor_results}
\end{figure}

In an experiment addressing scenarios with limited digital access to X-ray scans, such as rural areas in LMICs, we leveraged smartphones to directly capture images (Figure \ref{fig:monitor_results}), introducing specific artifacts such as Moiré effects and excessive contrast. We tested the Variance-Weighted Segmentation Loss (VWSL) on 5 samples and observed performance improvements of 5-20\% Dice. This outcome highlights our method's adaptability and efficiency, even with fewer samples, underscoring its practicality for real-world applications in resource-constrained settings by allowing for digitizing X-rays to assess diagnostics and monitor diseases remotely.


\subsection{Training on mobile devices}

To enable Neural Cellular Automata (NCA) model execution on smartphones, we ported Med-NCA to TensorFlow \cite{tensorflow2015-whitepaper} and subsequently to TensorFlow Lite, given its unique support for on-device training. This process was complemented by developing a specialized wrapper application, facilitating the loading and processing of both training and inference data on smartphones.

We assessed the training efficiency on five diverse smartphones: \emph{Moto G31}, \emph{Pixel 1 XL}, \emph{Poco M5}, \emph{Samsung S10}, and \emph{Pixel 6a}, chosen for their varied release years and price points. Training times varied, ranging from 38 to 82 hours, with the main influence being the release year and price point. Images of the app and exemplary results are available in the supplementary material.

All \emph{tested devices are used or refurbished}, enhancing their affordability and demonstrating the model's viability on cost-effective hardware. This approach emphasizes the potential for broad accessibility of advanced medical imaging technologies, especially in resource-constrained environments, by utilizing readily available and affordable smartphones.

\begin{figure}[t]
  \centering
  \includegraphics[width=.99\linewidth]{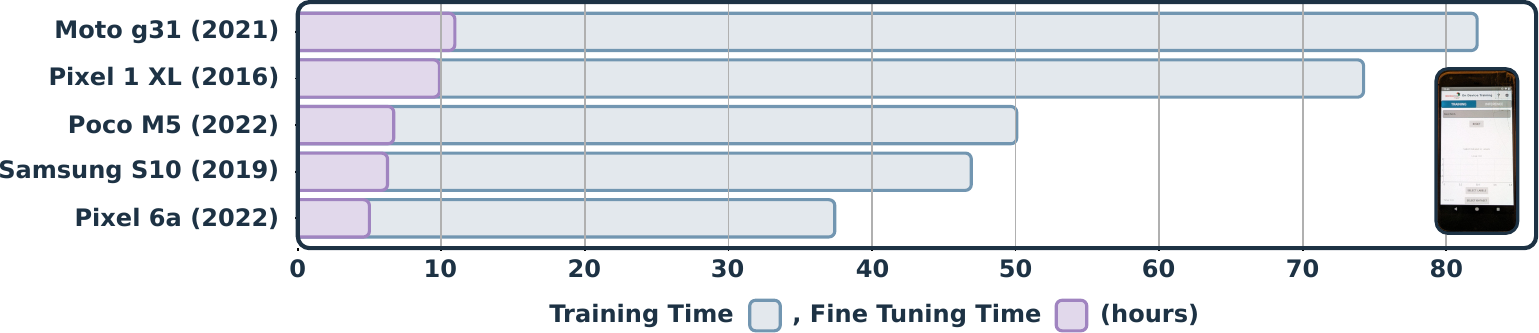}
  \caption{The training and finetuning time across different Android smartphones of different price classes and release years. With 50 training samples, full training is run for 1500 epochs and finetuning for 100 epochs.}
  \label{fig:training_times}
\end{figure}

\section{Conclusion}
Our study concludes an initial exploration of deploying Neural Cellular Automata (NCA) on smartphones for X-Ray segmentation, significantly advancing the accessibility of advanced diagnostic tools in regions with limited resources. By introducing the \emph{Variance-Weighted Segmentation Loss (VWSL) for unsupervised adaptation}, we have enabled efficient finetuning from unlabeled data, thereby improving model adaptability and performance across a spectrum of medical imaging contexts without the reliance on extensive computational resources or labeled datasets. Our evaluation across three multisite X-ray datasets—Padchest, ChestX-ray8, and MIMIC-III—for the segmentation of both lungs has not only demonstrated notable improvements in accuracy, with gains ranging from 0.7 to 2.8\% Dice but also underscored the practicality of implementing these sophisticated models on universally accessible smartphones. Direct \textbf{training and inference on smartphones}, tested across five devices, confirm the viability of advanced models for on-device use, greatly enhancing the accessibility of sophisticated medical diagnostics. 
In essence, this work aims to narrow the digital gap in medical imaging, ensuring that the transformative benefits of machine learning can reach healthcare practitioners and patients worldwide, regardless of their location or the resources at their disposal. Through the strategic deployment of NCAs and the introduction of our unsupervised adaptation method, we envision a future where personalized medical diagnostics become a global norm, contributing to the advancement of universal healthcare innovation.

\begin{credits}


\end{credits}

\bibliographystyle{splncs04}
\bibliography{Paper-1060}

\begin{thebibliography}{10}
\providecommand{\url}[1]{\texttt{#1}}
\providecommand{\urlprefix}{URL }
\providecommand{\doi}[1]{https://doi.org/#1}

\bibitem{tensorflow2015-whitepaper}
Abadi, M., Agarwal, A., Barham, P., Brevdo, E., Chen, Z., Citro, C., Corrado, G.S., Davis, A., Dean, J., Devin, M., Ghemawat, S., Goodfellow, I., Harp, A., Irving, G., Isard, M., Jia, Y., Jozefowicz, R., Kaiser, L., Kudlur, M., Levenberg, J., Man\'{e}, D., Monga, R., Moore, S., Murray, D., Olah, C., Schuster, M., Shlens, J., Steiner, B., Sutskever, I., Talwar, K., Tucker, P., Vanhoucke, V., Vasudevan, V., Vi\'{e}gas, F., Vinyals, O., Warden, P., Wattenberg, M., Wicke, M., Yu, Y., Zheng, X.: {TensorFlow}: Large-scale machine learning on heterogeneous systems (2015), \url{https://www.tensorflow.org/}, software available from tensorflow.org

\bibitem{adam2024medical}
Adam, D.: Medical ai could be'dangerous' for poorer nations, who warns. Nature  (2024)

\bibitem{resourceConstrained_overview}
Ajani, T.S., Imoize, A.L., Atayero, A.A.: An overview of machine learning within embedded and mobile devices--optimizations and applications. Sensors  \textbf{21}(13), ~4412 (2021)

\bibitem{bustos2020padchest}
Bustos, A., Pertusa, A., Salinas, J.M., De~La Iglesia-Vaya, M.: Padchest: A large chest x-ray image dataset with multi-label annotated reports. Medical image analysis  \textbf{66},  101797 (2020)

\bibitem{chen2021transunet}
Chen, J., Lu, Y., Yu, Q., Luo, X., Adeli, E., Wang, Y., Lu, L., Yuille, A.L., Zhou, Y.: Transunet: Transformers make strong encoders for medical image segmentation. arXiv preprint arXiv:2102.04306  (2021)

\bibitem{frija2021improve}
Frija, G., Bla{\v{z}}i{\'c}, I., Frush, D.P., Hierath, M., Kawooya, M., Donoso-Bach, L., Brklja{\v{c}}i{\'c}, B.: How to improve access to medical imaging in low-and middle-income countries? EClinicalMedicine  \textbf{38},  101034 (2021)

\bibitem{gaggion2023chexmask}
Gaggion, N., Mosquera, C., Mansilla, L., Aineseder, M., Milone, D.H., Ferrante, E.: Chexmask: a large-scale dataset of anatomical segmentation masks for multi-center chest x-ray images. arXiv preprint arXiv:2307.03293  (2023)

\bibitem{gilpin2019cellular}
Gilpin, W.: Cellular automata as convolutional neural networks. Physical Review E  \textbf{100}(3),  032402 (2019)

\bibitem{nnUnet}
Isensee, F., Jaeger, P.F., Kohl, S.A., Petersen, J., Maier-Hein, K.H.: nnu-net: a self-configuring method for deep learning-based biomedical image segmentation. Nature methods  \textbf{18}(2),  203--211 (2021)

\bibitem{johnson2016mimic}
Johnson, A.E., Pollard, T.J., Shen, L., Lehman, L.w.H., Feng, M., Ghassemi, M., Moody, B., Szolovits, P., Anthony~Celi, L., Mark, R.G.: Mimic-iii, a freely accessible critical care database. Scientific data  \textbf{3}(1), ~1--9 (2016)

\bibitem{kalkhof2023med}
Kalkhof, J., Gonz{\'a}lez, C., Mukhopadhyay, A.: Med-nca: Robust and lightweight segmentation with neural cellular automata. In: International Conference on Information Processing in Medical Imaging. pp. 705--716. Springer (2023)

\bibitem{kalkhof2023m3d}
Kalkhof, J., Mukhopadhyay, A.: M3d-nca: Robust 3d segmentation with built-in quality control. In: International Conference on Medical Image Computing and Computer-Assisted Intervention. pp. 169--178. Springer (2023)

\bibitem{lin2017focal}
Lin, T.Y., Goyal, P., Girshick, R., He, K., Doll{\'a}r, P.: Focal loss for dense object detection. In: Proceedings of the IEEE international conference on computer vision. pp. 2980--2988 (2017)

\bibitem{mccool2022mobile}
McCool, J., Dobson, R., Whittaker, R., Paton, C.: Mobile health (mhealth) in low-and middle-income countries. Annual Review of Public Health  \textbf{43},  525--539 (2022)

\bibitem{growingNCA}
Mordvintsev, A., Randazzo, E., Niklasson, E., Levin, M.: Growing neural cellular automata. Distill  \textbf{5}(2), ~e23 (2020)

\bibitem{paszke2019pytorch}
Paszke, A., Gross, S., Massa, F., Lerer, A., Bradbury, J., Chanan, G., Killeen, T., Lin, Z., Gimelshein, N., Antiga, L., et~al.: Pytorch: An imperative style, high-performance deep learning library. Advances in neural information processing systems  \textbf{32} (2019)

\bibitem{fernandoGarcia}
Perez-Garcia, F.: {fepegar/unet: First published version of PyTorch U-Net} (Oct 2019). \doi{10.5281/zenodo.3522306}, \url{https://doi.org/10.5281/zenodo.3522306}

\bibitem{UNet}
Ronneberger, O., Fischer, P., Brox, T.: U-net: Convolutional networks for biomedical image segmentation. In: International Conference on Medical image computing and computer-assisted intervention. pp. 234--241. Springer (2015)

\bibitem{vaswani2017attention}
Vaswani, A., Shazeer, N., Parmar, N., Uszkoreit, J., Jones, L., Gomez, A.N., Kaiser, {\L}., Polosukhin, I.: Attention is all you need. Advances in neural information processing systems  \textbf{30} (2017)

\bibitem{wang2017chestx}
Wang, X., Peng, Y., Lu, L., Lu, Z., Bagheri, M., Summers, R.M.: Chestx-ray8: Hospital-scale chest x-ray database and benchmarks on weakly-supervised classification and localization of common thorax diseases. In: Proceedings of the IEEE conference on computer vision and pattern recognition. pp. 2097--2106 (2017)

\end{thebibliography}

\begin{figure}[!htbp]
  \centering
  \includegraphics[width=.95\linewidth]{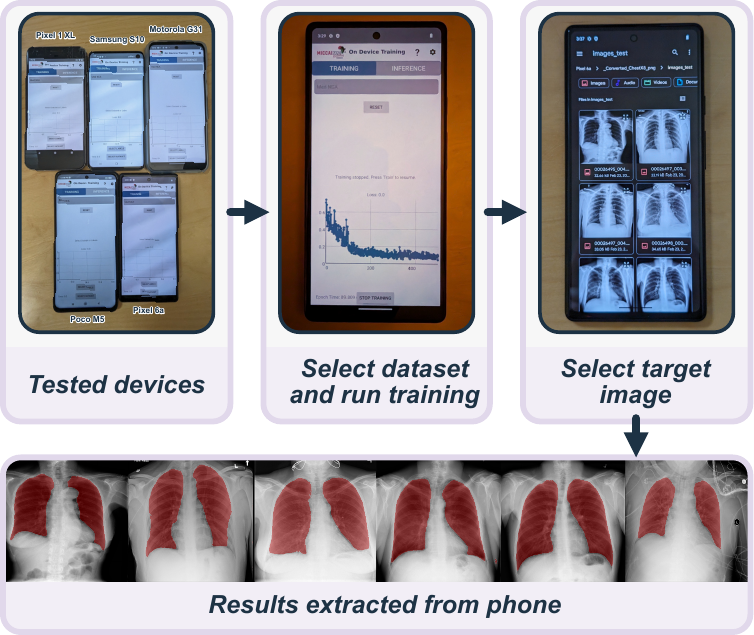}
  \caption{Visualization of the Android application for Med-NCA training and exemplary results of the inference.}
  \label{fig:TestSetupmobile}
\end{figure}

\end{document}